\documentclass[sigconf]{acmart}

\usepackage{booktabs} 

\copyrightyear{2017}
\acmYear{2017}
\setcopyright{acmcopyright} 
\acmConference{MM'17}{}{October 23--27, 2017, Mountain View, CA, USA.} 
\acmPrice{15.00} 
\acmDOI{https://doi.org/10.1145/3123266.3123279}
\acmISBN{978-1-4503-4906-2/17/10}

\usepackage{balance}
\usepackage{algorithm}
\usepackage{algorithmic}
\author{Longhui Wei$^{1}$, Shiliang Zhang$^{1}$, Hantao Yao$^{2,3}$, Wen Gao$^{1}$, Qi Tian$^{4}$}
\affiliation{
\small
  \institution{
 $^{1}$School of Electronic Engineering and Computer Science, Peking University, Beijing 100871, China \\
 $^{2}$Key Lab of Intelligent Information Processing of CAS, Institute of Computing Technology, CAS, Beijing 100190, China \\
 $^{3}$ University of Chinese Academy of Sciences, Beijing 100049, China \\
  $^{4}$ Department of Computer Science, University of Texas at San Antonio, San Antonio,  TX 78249-1604, USA \\}
}
\email{{longhuiwei,slzhang.jdl,wgao}@pku.edu.cn, yaohantao@ict.ac.cn, qitian@cs.utsa.edu}

\begin{document}
\title{GLAD: Global-Local-Alignment Descriptor for Pedestrian Retrieval}

\begin{abstract}
 The huge variance of human pose and the misalignment of detected human images significantly increase the difficulty of person Re-Identification (Re-ID). Moreover, efficient Re-ID systems are required to cope with the massive visual data being produced by video surveillance systems. Targeting to solve these problems, this work proposes a Global-Local-Alignment Descriptor (GLAD) and an efficient indexing and retrieval framework, respectively. GLAD explicitly leverages the local and global cues in human body to generate a discriminative and robust representation. It consists of part extraction and descriptor learning modules, where several part regions are first detected and then deep neural networks are designed for representation learning on both the local and global regions. A hierarchical indexing and retrieval framework is designed to eliminate the huge redundancy in the gallery set, and accelerate the online Re-ID procedure. Extensive experimental results show GLAD achieves competitive accuracy compared to the state-of-the-art methods. Our retrieval framework significantly accelerates the online Re-ID procedure without loss of accuracy. Therefore, this work has potential to work better on person Re-ID tasks in real scenarios.
\end{abstract}

%
%




\maketitle

\begin{figure}
\begin{center}
\includegraphics[width=1\linewidth]{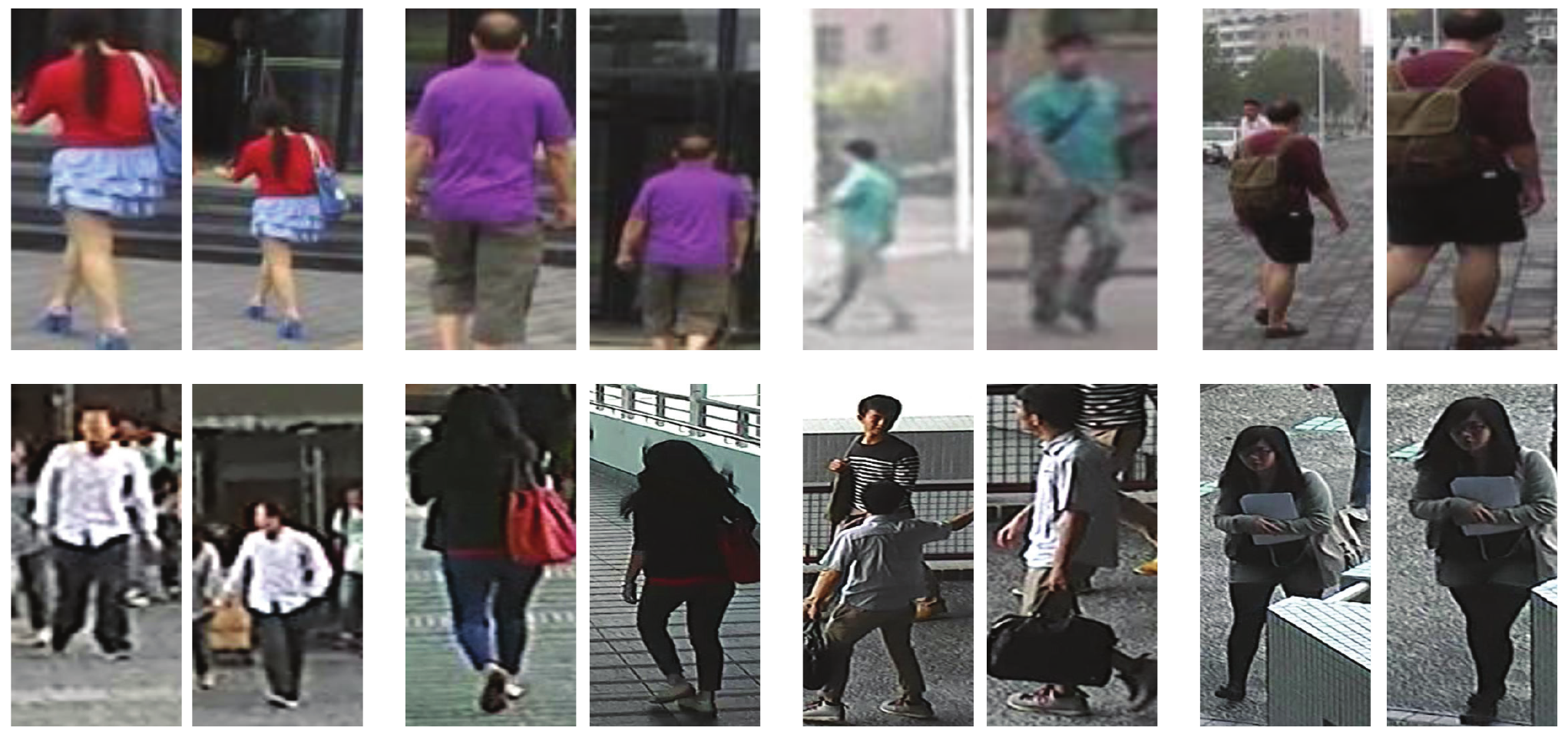}
\end{center}
\caption{Examples of detected pedestrian images from Market1501 (first row) and CUHK03 (second row).}
\label{fig:misAlignment}
\end{figure}

\section{Introduction}
\label{sec:intro}

Person Re-Identification (Re-ID) targets to probe and return images containing the identical query person from a gallery set. Because of its promising applications in video surveillance and public security, person Re-ID has drawn more and more attention in recent years. As shown in Fig.~\ref{fig:misAlignment}, the appearance of a person image can be easily affected by various factors like camera viewpoint, human pose, illumination, occlusion, \emph{etc}. Those make identifying a specific person from the large-scale gallery set a challenging task. To tackle this challenge, most of person Re-ID works focus on two stages, \emph{i.e.}, descriptor learning and distance metric learning. Descriptor learning aims to learn a discriminative descriptor to represent the appearances of different persons. Distance metric learning is designed to reduce the distance among descriptors of images containing the same person. Traditional descriptor learning methods usually extract rigid local invariant features. Suffering from the huge variance of human pose and camera viewpoint, these descriptors are not robust enough and often fail to identify person. Most of distance metric learning methods take a pair of pedestrian images as input, thus correspond to the high complexity. Inspired by the success of Convolutional Neural Networks (CNN) in large-scale visual classification, latest works start to design deep learning algorithms and have achieved significant improvements.

Most of deep learning based works learn descriptors from the whole pedestrian images. Such descriptors thus depict the global cues but may lose crucial details. This problem has been noticed in previous works~\cite{yi2014deep,wu2016enhanced,cheng2016person,hantao}, where the researchers divide the whole pedestrian image into several fixed-length strips. By learning descriptors on these strips rather than the whole image, these methods~\cite{yi2014deep,wu2016enhanced,cheng2016person,hantao} explicitly embed more detailed local cues. Experiments show such descriptors substantially outperform the global descriptors. However, fixed-length strips would be sensitive to the pose variance and person misalignment. As illustrated in Fig. \ref{fig:misAlignment}, misalignment commonly exists in detected pedestrian images. Therefore, more delicate ways should be designed to overcome the pose variance and misalignment issues.

Aiming to solve the above issues, we propose the Global-Local-Alignment Descriptor (GLAD), which is generated with two modules,~\emph{i.e.,} part extraction and descriptor learning. Part extraction module utilizes Deeper Cut~\cite{DeeperCut} to estimate four human key points that are robust for various poses and camera viewpoints. Three coarse part regions, \emph{i.e.,} head, upper-body, and lower-body are hence generated based on the estimated key points. To explicitly embed part cues in the learned representation, a CNN composed of four sub-networks is proposed in descriptor learning module. Those sub-networks share several convolutional layers and are designed to learn descriptors on three part regions and the global image, respectively. During the training stage, the shared convolution layers can be efficiently optimized by multiple learning tasks on different body regions to avoid overfitting. After network training, we feed the three part regions and global image into the neural network to extract four descriptors, which are finally concatenated as the GLAD. Therefore, GLAD contains both the global and local cues, thus is potential to be more discriminative. GLAD also could be more robust to the pose variance and image misalignment issues because the human body is divided in a more meaningful way.

Person Re-ID is commonly solved as a classification or distance metric learning problem~\cite{yi2014deep, wu2016personnet,ahmed2015improved,WARCA, LOMOXQAD, LDNS, Su3, Su4}, which suffer from massive computations of classifier training and the high time complexity for online pairwise matching. To make our Re-ID system scalable on large-scale datasets, we regard person Re-ID as a fine-grained pedestrian retrieval task, and focus on designing a more efficient indexing and retrieval framework. In person Re-ID gallery sets, each person would has multiple samples. This implies data redundancy and could be optimized by indexing strategies. Our indexing algorithm is thus designed to group samples of the same person into one unit. Specifically, we propose a Two-fold Divisive Clustering algorithm (TDC) to group different samples of one person together through dividing samples of different persons in a greedy manner. Finally, a descriptor is generated to depict the visual cues of each group. The online Re-ID procedure can be regarded as a two-folder retrieval, where the coarse retrieval retrieves image groups, and a fine retrieval is then conducted to get a precise image ranklist. In other words, we need not match the query person image against each gallery image during retrieval procedure. So, our retrieval strategy can effectively speed up the online Re-ID.

Although there are many deep learning based person Re-ID works, our work differs from them in the aspects of introducing a more efficient online Re-ID strategy and considering delicate part cues. Zheng~\emph{et al.}~\cite{zheng2017pose} also propose a pose invariant embedding framework to solve the misalignment issue. Ten fine-grained parts are extracted by estimating human key points. These parts are first normalized by affine transformation, then are combined to compose a global pose invariant image. The final representation is hence extracted from the standard pose image. Therefore, the representation in~\cite{zheng2017pose} is not learned explicitly on local parts and still belongs to the global representation. Moreover, as shown in our experiments, fine-grained part extraction is easily affected by image noises, pose and viewpoint variances. For instance, arms can be invisible due to occlusion or pose changes. Our experiments also show mandatory detection of fine-grained parts, \emph{e.g.}, arms, results in noisy part regions and degrades the Re-ID performance. Extensive experimental results on three public datasets show our GLAD and retrieval framework present competitive accuracy and efficiency compared to the state-of-the-art methods. Our method also presents substantial advantages on automatically detected pedestrian images. Therefore, we conclude this work has potential to be more robust and effective in real scenarios, and our contribution is valuable.

\section{Related Work}\label{sec:related}

This work is related with deep learning based person Re-ID and human part detection for person Re-ID. The following parts briefly review several works on these two categories, respectively.

\subsection{Deep Learning based Person Re-ID}

Deep learning shows remarkable performance in computer vision and multimedia tasks and has become the main stream method for person Re-ID. Current deep learning based person Re-ID methods can be divided into two categories based on the usage of deep neural network, \emph{i.e.}, feature learning and distance metric learning. Feature learning networks aim to learn a robust and discriminative feature to represent pedestrian images. Cheng~\emph{et al.}~\cite{cheng2016person} propose a multi-channel parts based network to learn a discriminative feature with an improved triplet loss. Wu~\emph{et al.}~\cite{wu2016enhanced} discover hand-crafted feature is complementary with CNN feature. They thus divide one image into five fixed-length part regions. For each part region, a histogram descriptor is generated and concatenated with the full body CNN feature. Su~\emph{et al.}~\cite{su2016deep,Su2} propose a semi-supervised attribute learning framework to learn binary attribute features. In~\cite{zheng2016discriminatively}, identification model and verification model are combined to learn a discriminative representation. In \cite{xiao2016learning}, a new dropout algorithm is designed for feature learning on a multi-domain dataset, which is generated by combining several existing datasets.

Siamese network is commonly used to learn better distance metrics between the input image pair. Yi~\emph{et al.}~\cite{yi2014deep} propose a siamese network composed of three components,~\emph{i.e.}, CNN, connection function, and cost function, respectively. Similar with~\cite{cheng2016person}, several fixed-length part regions are divided and trained independently. In~\cite{wu2016personnet}, an end-to-end siamese network is proposed. By utilizing small filters, the network goes deeper and obtains a remarkable performance. Ahmed~\emph{et al.}~\cite{ahmed2015improved} design a new layer to capture local relationships between input image pair. In~\cite{liu2016end}, comparative attention network is proposed to adaptively compare the similarity between images.

\subsection{Human Part Detection for Person Re-ID}

Human parts provide important local cues of human appearance. Therefore, it is natural to design part detection algorithms for person Re-ID in some early person Re-ID works~\cite{farenzena2010person, cheng2014person, bedagkar2012part}. Motivated by the symmetry and asymmetry properties of human body, Farenzena~\emph{et al.}~\cite{farenzena2010person} propose to detect salient part regions by the perceptual principles of symmetry and asymmetry. In~\cite{cheng2014person}, Cheng~\emph{et al.} propose a pictorial structure algorithm to detect parts. In~\cite{bedagkar2012part}, deformable part model~\cite{f2008} is utilized to detect six body parts. Most of recent deep learning based methods directly divide pedestrian images into fixed-length regions and have not paid much attention in leveraging part cues~\cite{li2014deepreid}. Recently, Zheng~\emph{et al.}~\cite{zheng2017pose} adopt the convolution pose machines~\cite{wei2016convolutional} to detect fine-grained body parts and then generate a standard pose image, which is hence utilized to generate descriptors. Therefore, the representation~\cite{zheng2017pose} is not learned explicitly on local parts. Also, fine-grained part extraction is expensive and could be easily affected by image noises, pose and viewpoint variance. Those factors would degrade the Re-ID accuracy and efficiency.

\section{Problem Formulation}

Given a probe image $p$, person Re-ID targets to identify and return images containing the identical person in $p$ from a set of gallery images $\{(g_{1},l_{1}),(g_{2},l_{2}),...,(g_{N},l_{N})\}$, where $g_{i}$ and $l_{i}$ denote the $i$-th gallery image and its person ID label, respectively. Person Re-ID can be tackled by classifying those gallery images~\cite{zheng2016discriminatively, wu2016personnet, wu2016enhanced, su2016deep}, or by an image retrieval procedure, \emph{i.e.}, ranking those images based on a descriptor and a distance metric $d(\mathbf{f}_{p}, \mathbf{f}_{i})$, where $\mathbf{f}$ represents the generated image descriptor, and $d(\cdot)$ denotes the distance between probe image and gallery image. The returned ranklist of $N$ images could be denoted as $\{r_{1},r_{2},...,r_{N}\}$, where $r_{i}$ is the sorted index of image $g_{i}$. Under the retrieval formulation, the objective function of person Re-ID can be summarized as Eq.~\eqref{Eq:formulation},\emph{ i.e.},
\begin{equation}
\min\sum_{i=1}^{N} r_{i} \textbf{I}(l_{p},l_{i}),\qquad \textbf{I}(l_{p},l_{i})=\left \{ \begin{array}{ll} 1 & l_{p}=l_{i} \\ 0 & l_{p}\not= l_{i} \end{array} ,\right.
\label{Eq:formulation}
\end{equation}
where $l_{p}$ is the person ID label of the probe image $p$.

Compared with person classification, treating person Re-ID as a retrieval task has potential to better cope with large-scale data and present improved generalization ability to unseen samples. Therefore, the retrieval formulation may work better in real scenarios, because the probe persons commonly do not exist in the training set. Under the retrieval formulation, person Re-ID consists of two critical steps: 1) robust and discriminative descriptor generation, and 2) efficient image similarity computation and ranking.


Targeting to deal with the image misalignment and pose variance issues, we present Global-Local-Alignment Descriptor in Sec.~\ref{sec:GLN}. Most of previous Re-ID works focus on descriptor generation, and has not paid much attention to efficient gallery image indexing and ranking. In Sec. \ref{sec:retrieval}, we propose an efficient indexing and retrieval framework that makes person Re-ID using GLAD more efficient.

\section{Global-Local-Alignment Descriptor} \label{sec:GLN}

\begin{figure}
\begin{center}
\includegraphics[width=1\linewidth]{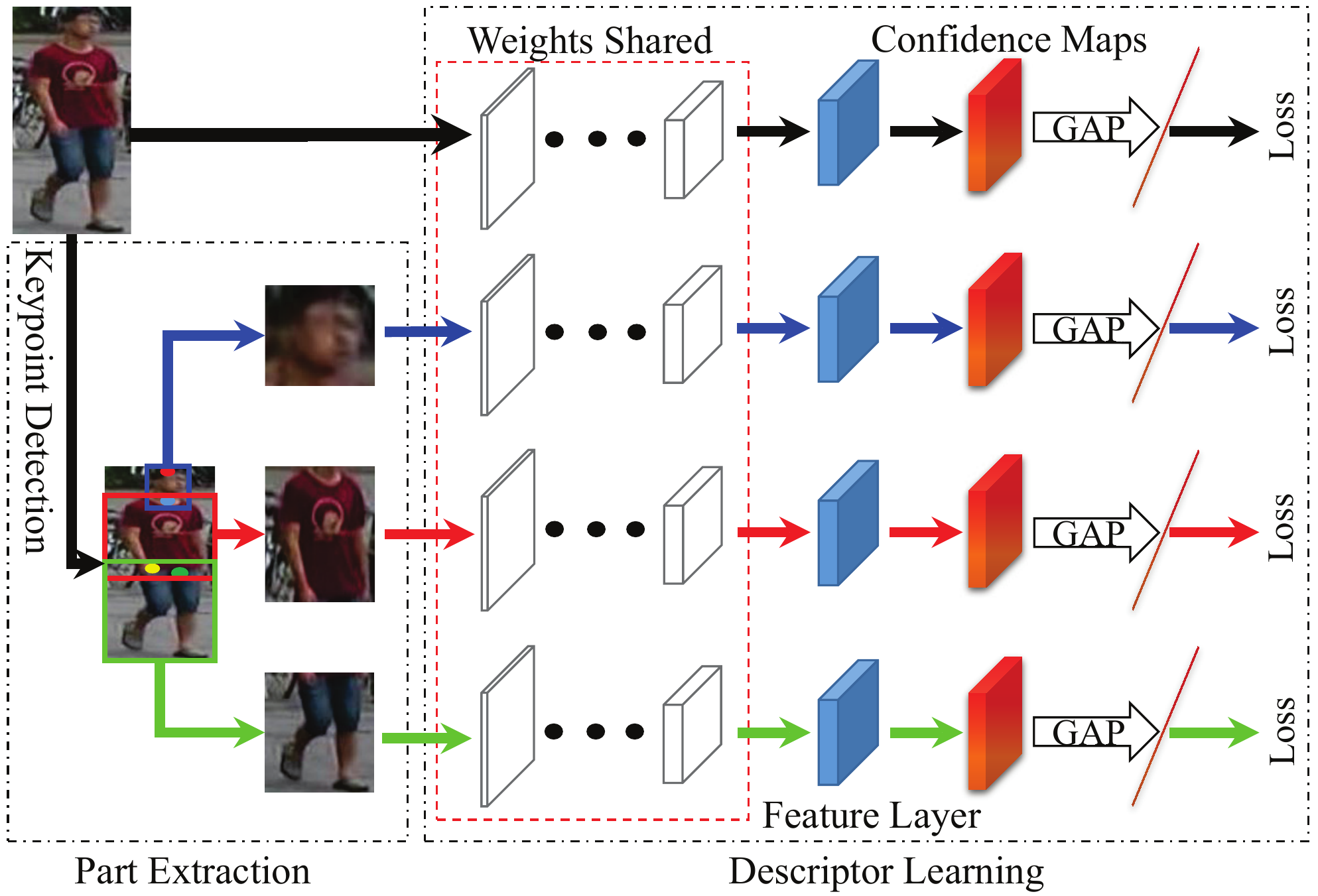}
\end{center}
\caption{Framework of GLAD extraction, which includes two modules,~\emph{i.e.}, part extraction and descriptor learning. Three parts are extracted based on four detected key points. A four-stream CNN is designed to generate descriptors from both the global and local regions.}
\label{fig:structure}
\end{figure}

The framework of GLAD extraction is summarized in Fig.~\ref{fig:structure}. It can be observed that, we first detect several body parts from an input person image, then learn descriptors from both the global and local regions. Through detecting more subtle body parts, GLAD has potential to be robust to the misalignment and would gain more discriminative power by explicitly embedding global and local cues. In the following, we present the part extraction module and descriptor learning module, respectively.

\subsection{Part Extraction}

Body part extraction has been studied by many pose estimation works~\cite{pose1, pose2, pose3, DeepCut, DeeperCut}. However, the pedestrian images in person Re-ID are taken in unconstrained environment, and are easily affected by occlusions, viewpoint changes, and pose variances. Those factors make it difficult to detect fine-grained parts. For example, either the left or right arms cannot be detected in side view images of pedestrian. Mandatory detection of such parts results in noisy part regions and may degrade the Re-ID performance. The above issues motivate us to consider parts that could be easily and reliably detected under various viewpoint and pose changes.

\begin{figure}
\begin{center}
\includegraphics[width=0.9\linewidth]{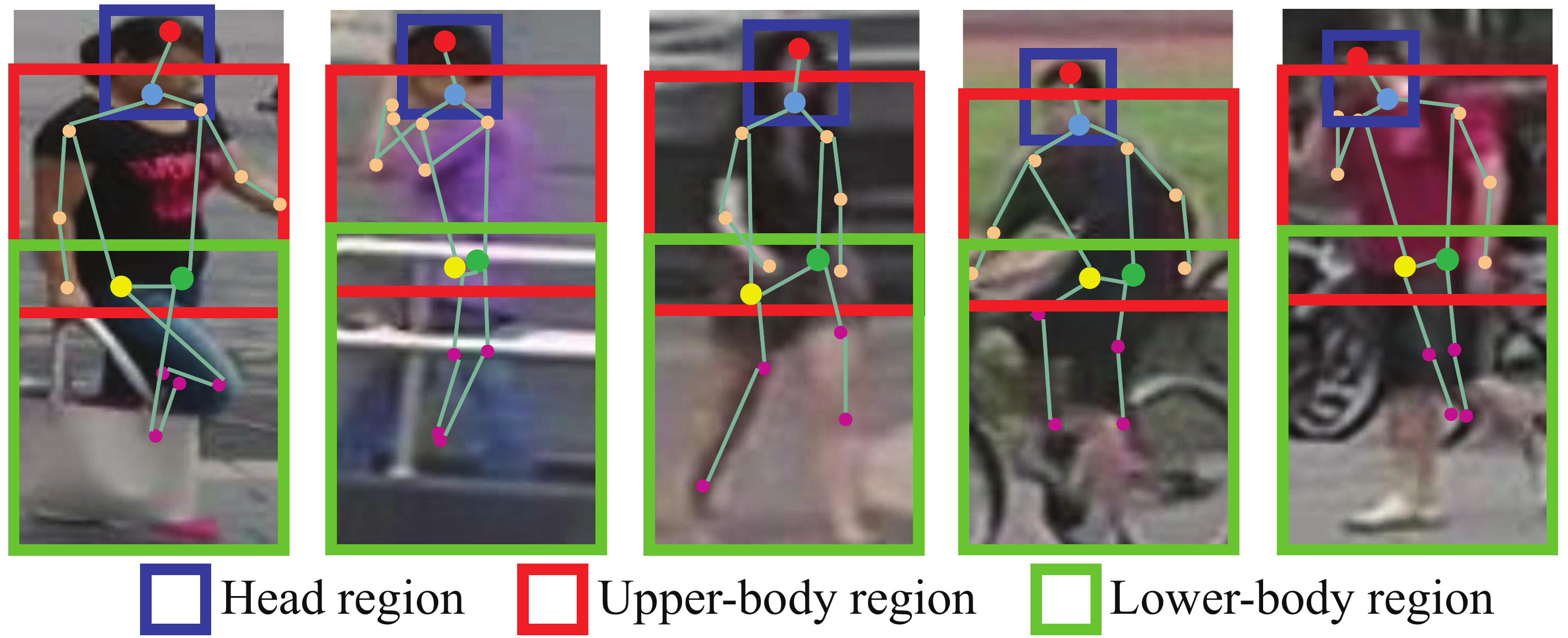}
\end{center}
\caption{Examples of detected keypoints and generated three part regions. The four keypoints used in our method are emphasized with large size.}
\label{fig:part_extraction}
\end{figure}

Specifically, we utilize Deeper Cut~\cite{DeeperCut} to estimate only four key points,~\emph{i.e.,} upper-head, neck, right-hip, left-hip, respectively on the pedestrian image. As shown in Fig.~\ref{fig:part_extraction}, based on those four key points, we can coarsely divide a pedestrian image into three part regions: head, upper-body, and lower-body, respectively. The head region can be located based on upper-head point and neck point. Suppose the size of person image is $H \times W$ and the coordinates of upper-head point and neck point are $(x_1,y_1)$ and $(x_2,y_2)$, we crop the head region $B^h$ with Eq. \eqref{eq:head}, \emph{i.e.},
\begin{equation}
\begin{aligned}
&B^h = [(x_c-w/2, y_1-\alpha), (x_c+w/2, y_2+\alpha)], \\
&w =  y_2-y_1+2\cdot\alpha,  \\
&x_c = (x_1+x_2)/2,
\label{eq:head}
\end{aligned}
\end{equation}
where the $B^h$ is located by coordinates of the upper-left and bottom-right points. $\alpha$ is a parameter controlling the overlap between neighboring parts regions. $\alpha$ is experimentally set as 15 for the $512 \times 256$ sized person image.

Suppose the coordinates of left-hip and right-hip points are $(x_3,y_3)$ and $(x_4,y_4)$, the upper-body region $B^{ub}$ and the lower-body region $B^{lb}$ can be captured in similar way with Eq. \eqref{eq:UL}, \emph{i.e.},
\begin{equation}
\begin{aligned}
&B^{ub} = [(0, y_2-2\cdot\alpha), (W-1, y_c+2\cdot\alpha)], \\
&B^{lb} = [(0, y_c-2\cdot\alpha), (W-1, H-1)], \\
&y_c = (y_3+y_4)/2,
\end{aligned}
\label{eq:UL}
\end{equation}

Examples of detected keypoints and part regions are illustrated in Fig.~\ref{fig:part_extraction}. It can be observed that, the four key points,~\emph{i.e.,} upper-head, neck, right-hip, left-hip are more robust to pose and viewpoint changes than the other keypoints. The three part regions hence could be reliably extracted. Because the keypoints on human foot are not stable, it is difficult to confirm the bottom coordinate of lower-body region. We thus simply set the bottom of the image as the bottom of lower-body region. More extensive evaluations on the validity of part extraction will be given in Sec.~\ref{sec:effect_of_part}.

\subsection{Descriptor Learning}

Existing deep neural networks such as~\emph{AlexNet} \cite{alexnet}, \emph{GoogLeNet} \cite{googlenet}, \emph{VGGNet} \cite{VGGnet}, and~\emph{ResNet}~\cite{resnet} have been utilized to learn descriptors on
the global image for person Re-ID. To explicitly leverage global and local cues for descriptor learning,
we propose a four-stream CNN. As illustrated in Fig.~\ref{fig:structure}, the proposed network includes one sub-network for global descriptor learning and three sub-networks for part descriptor learning, respectively. These sub-networks share the identical structure and can be initialized by exiting network structures and parameters.

Specifically, our network is modified and initialized from \emph{GoogLe-}
 \emph{Net} \cite{googlenet} by replacing its fully connected layers with two convolutional layers as classifier. As shown in Fig.~\ref{fig:structure}, we call the first convolutional layer as feature layer because it is used for feature extraction. The latter convolutional layer directly produces $C$ feature maps corresponding to $C$ classes in the training set. Therefore, we call those feature maps as confidence maps, which essentially show the classification confidences.

Based on the confidence maps, we apply Global Average Pooling (GAP) to generate the classification score for each class. GAP averages the responses on each two-dimensional feature map, \emph{i.e.},
\begin{eqnarray}
S_c = \frac{1}{X \times Y}\sum_{x=1}^{X}\sum_{y=1}^{Y}M_{c}(x,y),
\end{eqnarray}
where $S_c$ denotes the classification score of the $c$-th class, and $M_{c}(x,y)$ is the response value at location of $(x,y)$ on the confidence map corresponding to the $c$-th class. $X$ and $Y$ are the width and height of the confidence map, respectively. Following GAP, softmax loss function is used to 
compute the network loss.

This updated architecture removes the fully connected layers and shows several advantages for feature learning. 1) It has fewer parameters, thus could better avoid overfitting on small training sets. 2) Without fully connected layers, it accepts images with arbitrary scales as input. We thus could resize the input image into larger scales to allow the neural network capture more detailed cues. Experiment result shows our network generates more discriminative feature than many existing algorithms.

For the global descriptor learning, the input image is the original image with scale resized to 512 $\times$ 256. For descriptor learning on head, the head region is resized to 96 $\times$ 96 as the network input. For upper-body and lower-body sub-networks, the input size is set as 224 $\times$ 256, respectively. These sub-networks are trained in different classification tasks, \emph{i.e.}, each task aims to classify the global or local input regions into correct person classes. As illustrated in Fig.~\ref{fig:structure}, instead of training the four sub-networks alone, we train them together with sharing weights in convolution layers. This optimizes the convolutional layers in different tasks and hence better avoids overfitting. We evaluate this strategy in Sec.~\ref{sec:GLD_B}.

During testing, we use the feature maps produced by feature layer to generate descriptors. Suppose $M$ channels of feature maps are generated in the feature layer, we finally generate an $M$ dimensional feature descriptor by GAP on each feature map. The descriptors extracted on four regions are concatenated as the final GLAD, \emph{i.e.},
\begin{eqnarray}
\textbf{f}^{GLAD} = [ \textbf{f}^{G}; \textbf{f}^{h}; \textbf{f}^{ub}; \textbf{f}^{lb} ],
\label{eq:Global-Local-D}
\end{eqnarray}
where $\textbf{f}^{GLAD}$ denotes the final GLAD. $\textbf{f}^{G}$ represents the learned feature on the global image, $\textbf{f}^{h}$, $\textbf{f}^{ub}$ and $\textbf{f}^{lb}$ are descriptors generated from the three sub-networks, respectively. Therefore, GLAD is an $4 \times M$ dimensional vector, which explicitly contains global and local cues. We experimentally set $M$ as 1024, which generates an 4096-dim GLAD.

By only detecting robust coarse part regions, GLAD seeks a reasonable trade-off between part detection accuracy and robustness to misalignment and pose changes. Therefore, GLAD would be more robust to misalignment issues than global features. Moreover, GLAD is trained with multiple losses computed on different regions. This essentially enforces the network to focus on different parts and learn discriminative feature for each of them. This training strategy has potential to learn more discriminative features than previous deep features, which may get overfitted to the most discriminative parts on the training set and ignores the others. Detailed evaluations on GLAD will be presented in Sec. \ref{sec:GLD_B}.

\section{Retrieval Framework}
\label{sec:retrieval}

\begin{figure}
\begin{center}
\includegraphics[width=1\linewidth]{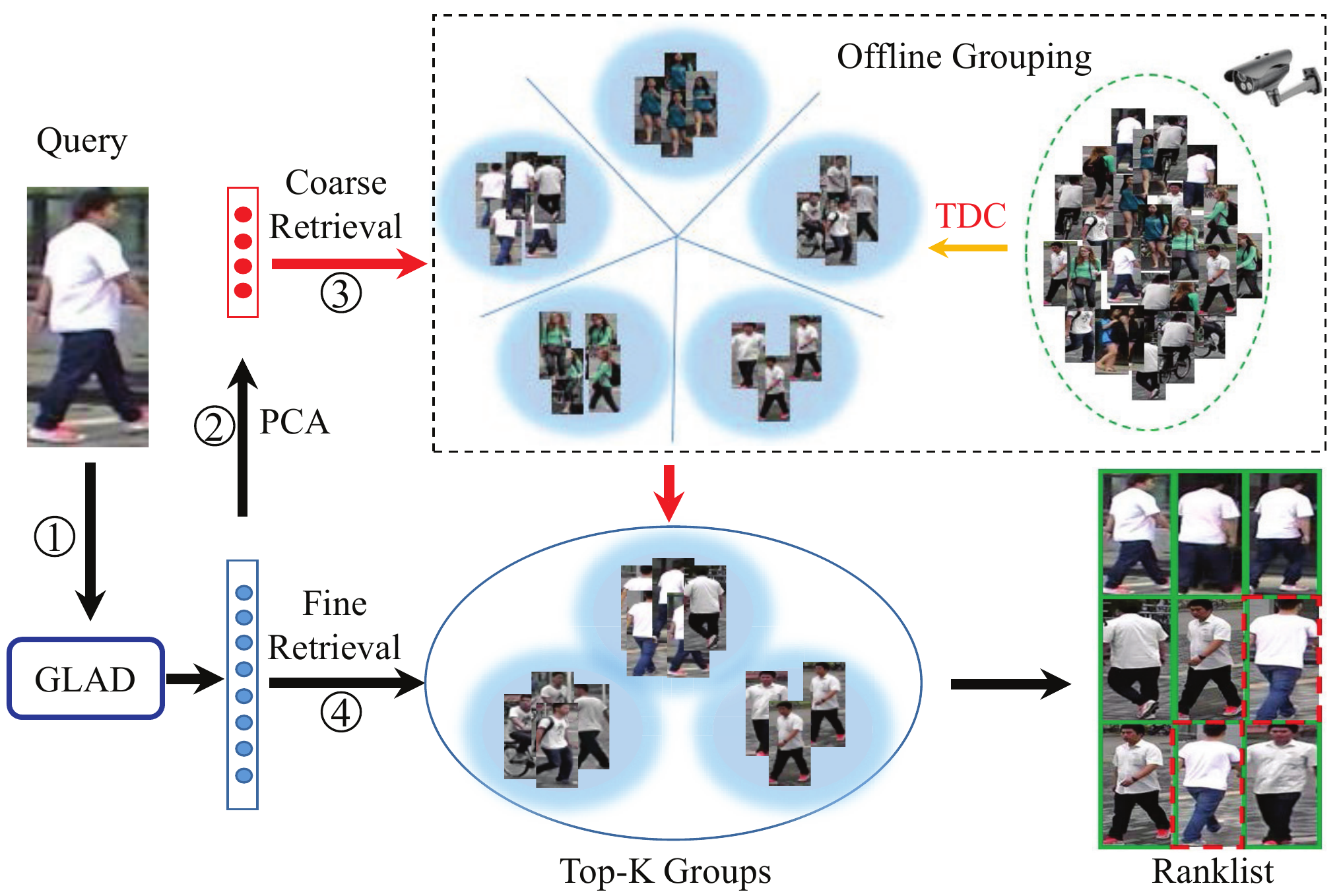}
\end{center}
\caption{Our retrieval framework mainly contains two modules,\emph{ i.e.}, offline grouping and coarse-to-fine online retrieval. Two-fold Divisive Clustering (TDC) gathers similar images into groups. Images in returned groups are retrieved with original GLAD to generate an image ranklist.}
\label{fig:offline}
\end{figure}

Based on the GLAD, we proceed to propose a hierarchical indexing and retrieval framework illustrated in Fig.~\ref{fig:offline}. As shown in Fig. \ref{fig:offline}, the offline indexing stage clusters similar images into the same group. This is motivated by the fact that, each person has multiple samples in the gallery set, \emph{e.g.}, one person can be recorded for multiple times by different cameras. Carefully grouping these samples together thus reduces the data redundancy and improves the online retrieval efficiency. Moreover, sample grouping is potential to significantly improve the accuracy of person Re-ID. For each person in the gallery set, the generated group may contain both his/her samples that can be easily identified and samples that can be hardly identified. Those hard samples thus can be retrieved together with the easy samples as one group during online retrieval. Therefore, efficient offline grouping algorithms should be designed.

There are many ways to cluster the images into groups \cite{clustering_Zhang}. Because the number of identities in person Re-ID gallery is unknown, it is hard to set the group number manually and makes clustering methods like K-Means \cite{kmeans} not optimal for this task. We thus propose a clustering method, called Two-fold Divisive Clustering (TDC) that does not need to manually specify the group number. Similar with H-LDC \cite{MTM}, TDC is a greedy strategy that divides images in galley into groups and ensures samples in each group share strong similarity with each other. For TDC, the group dissimilarity degree measurement is defined as
\begin{eqnarray}
D^{dis} = \frac{1}{N \times (N-1)}\sum_{i=1}^N\sum_{j=1}^N dis(g_i,g_j),
\label{eq:Measurement}
\end{eqnarray}
where $D^{dis}$ denotes the dissimilarity degree within a group, and $N$ is the number of images in the group. $dis(\cdot)$ represents the squared Euclidean distance between two images in the group, \emph{i.e.}, $g_i$ and $g_j$.

TDC is conducted to divide the image gallery into groups in a greedy manner, and finally ensures the dissimilarity degree within each group below a threshold $\theta$. Details of TDC are summarized in Algorithm \ref{alg:TDC}. Compared with K-Means related methods, TDC does not require the pre-defined cluster number and only has one parameter, \emph{i.e.}, the dissimilarity degree threshold $\theta$. Our experiments in Sec.~\ref{sec:E_RS} show that this parameter could be easily tuned. Our current work computes $dis(g_i,g_j)$ with GLAD. More discriminative features could be leveraged to further improve the quality of groups. This will be studied in our future work.

\label{sec:off}
\begin{algorithm}[htb]
\caption{Two-fold Divisive Clustering}
\label{alg:TDC}
\begin{algorithmic}[1]

	\REQUIRE Gallery $ \{ g_1,g_2,...,g_N\}$, dissimilarity threshold $\theta$
	 \ENSURE Group set $\mathcal{R}$
	\STATE \textbf{Initialization:} $\mathcal{G}_{1} = \{ g_1,g_2,...,g_N\}$, $\mathcal{R} = \{\mathcal{G}_1\}$,  $Count = 1$	
    \STATE \WHILE {True}
   \STATE $\forall \mathcal{G}_{*} \in \mathcal{R}$: compute $D^{dis}_{*}$ according to Eq. \eqref{eq:Measurement}\
   \IF {$\exists D^{dis}_{*}> \theta$}
    \FOR {each $\mathcal{G}_*$ with $D^{dis}_* > \theta$}
    \STATE  Choose the furthest two samples $(g_l,g_r)$ in $\mathcal{G}_*$
    \STATE  $\mathcal{G}_{Count+1}=\{\}$, $\mathcal{G}_{Count+2}=\{\}$
    \FOR { $g_j \in \mathcal{G}_*$}
    \IF {$dis(g_j,g_l) < dis(g_j,g_r)$}
    \STATE   $\mathcal{G}_{Count+1} \leftarrow  \mathcal{G}_{Count+1} \cup g_j$
    \ELSE
    \STATE  $\mathcal{G}_{Count+2} \leftarrow  \mathcal{G}_{Count+2} \cup g_j$
    \ENDIF
    \ENDFOR

    \STATE  $\mathcal{R} \leftarrow  \mathcal{R} \cup \mathcal{G}_{Count+1} \cup \mathcal{G}_{Count+2}$
    \STATE  $\mathcal{R} \leftarrow  \mathcal{R} \setminus \mathcal{G}_*$
    \STATE  $Count = Count + 2$
    \ENDFOR
        \ELSE
    \STATE Break
    \ENDIF

    \ENDWHILE

\end{algorithmic}
\end{algorithm}

After offline clustering images into groups, we generate a group descriptor to depict the visual appearance of each group. We generate the group descriptor wth Eq. \eqref{eq:GroupFeature}, \emph{i.e.},

\begin{eqnarray}
\mathbf{f}^{\mathcal{G}}(i) = \frac{1}{N}\sum_{j=1}^N \mathbf{f}_{j}^{GLAD}(i),
\label{eq:GroupFeature}
\end{eqnarray}
where $\mathbf{f}^{\mathcal{G}}(i)$ denotes the $i$-th dimension of group descriptor $\mathbf{f}^{\mathcal{G}}$. $N$ is the number of samples in the group, and $\mathbf{f}_{j}^{GLAD}(i)$ is the $i$-th dimension in GLAD of the ${j}$-th sample. For every group, we can get an 4096-dim feature descriptor. To speed up the similarity computation, we reduce the dimensionality of $\mathbf{f}^{\mathcal{G}}$ into 128 with PCA for fast online retrieval.

As shown in Fig.~\ref{fig:offline}, the online retrieval first retrieves image groups. GLAD is first extracted from the query, then is converted into an 128-dim vector with PCA. The 128-dim feature is used to retrieve relevant image groups. Because the number of image groups is significantly smaller than the number of images, this procedure can be efficiently finished. After the coarse retrieval, the top $K$ relevant groups are selected for fine retrieval, \emph{i.e.}, the original 4096-dim GLAD is used to rank the images contained in the $K$ groups to get a precise image ranklist. We experimentally set $K$ as 100. The two stages are performed to first quickly narrow-down the search space, then refine the initial result, respectively. Thus, they are combined to improve both the Re-ID efficiency and accuracy. This retrieval framework is evaluated in Sec. \ref{sec:E_RS}.

\section{Experiments}
\label{sec:exper}

\subsection{Datasets}\label{sec:dataset}

We evaluate the proposed methods on three widely used person Re-ID datasets,~\emph{i.e.}, \emph{Market1501}~\cite{market1501}, \emph{CUHK03}~\cite{li2014deepreid}, and \emph{VIPeR}~\cite{viper}.

Market1501~\cite{market1501} is composed of 1,501 identities automatically detected from six cameras. The dataset clearly defines and splits training and testing sets. The training set contains 12,936 images of 751 identities. 19,732 images of 750 identities are included in the testing set. Market1501 is a large-scale dataset and is designed for pedestrian retrieval task. Therefore, mean Average Precision (mAP) is also used to evaluate person Re-ID algorithms.

CUHK03~\cite{li2014deepreid} consists of 1,467 identities captured from two cameras. Automatically detected images by pedestrian detector and human labeled bounding boxes are both provided. On average, each person has 4.8 images under each camera. CUHK provides 20 split sets, each randomly selects 1,367 identities for training and the rest fort testing. We choose the first split set and report the average accuracy after repeating the experiments for 1,000 times.

VIPeR~\cite{viper} is smaller than Market1501 and CUHK03. It contains 632 identities and 1,264 images taken by two cameras. 316 identities are randomly chosen as training data, the rests are hence used as testing data. Because of its small size, the training set in VIPeR is enhanced to make deep model training possible. More details of training procedure on VIPeR are summarized in Sec. \ref{sec:Train}.

\subsection{Implementation Details}\label{sec:Train}

\begin{table}
\centering
\small
\caption{Comparison of different feature fusion and training strategies on Market1501. Baseline denotes the descriptor generated by our modified GoogLeNet \cite{googlenet} on the original image.}
\begin{tabular}{lcccc}
\hline
 Training Strategy & Descriptor  & mAP &  Rank-1\\
\hline
- &Baseline    & $60.3$  & $80.7$ \\
\hline
& Global & $60.3$  & $80.7$ \\
& Upper+Lower body     & $53.8$ & $79.8$ \\
WO/S& Head+Upper+Lower body       & $49.6$ & $77.3$ \\
& Head+Upper+Lower body (W)     & $55.7$ & $81.0$ \\
& GLAD     & $\bf{71.0}$ & $\bf{87.9}$\\
\hline
& Global & $66.1$  & $84.6$ \\
& Upper+Lower body    & $60.9$ & $84.2$ \\
W/S& Head+Upper+Lower body       & $55.6$ & $81.8$ \\
& Head+Upper+Lower body (W)    & $62.8$ & $85.5$ \\
& GLAD    & $\bf{73.9}$ & $\bf{89.9}$\\
\hline
\end{tabular}
\label{tab:GLN_Combination}
\end{table}

We use Caffe~\cite{caffe} to implement the neural networks. To estimate keypoints for GLAD extraction, we use Deeper Cut~\cite{DeeperCut} model pre-trained on the MPII human pose dataset~\cite{MP}. During GLAD learning procedure, an initial learning rate is set as 0.001, and is divided by 2 after every 20,000 iterations. Fine-tuning is applied on the target training set to avoid overfitting. On Market1501 and CUHK03, we train our network with 100,000 iterations. On VIPeR, we combine VIPeR training set with the training sets of CUHK03 and Market1501, then train the neural network on this mixed dataset with 100,000 iterations. All experiments are conducted on a server equipped with GeForce GTX 1080 GPU, Intel i7 CPU, and 32GB memory.

\subsection{ Evaluation on Descriptor Learning}
\label{sec:GLD_B}
 
\begin{figure}
\begin{center}
\includegraphics[width=0.85\linewidth]{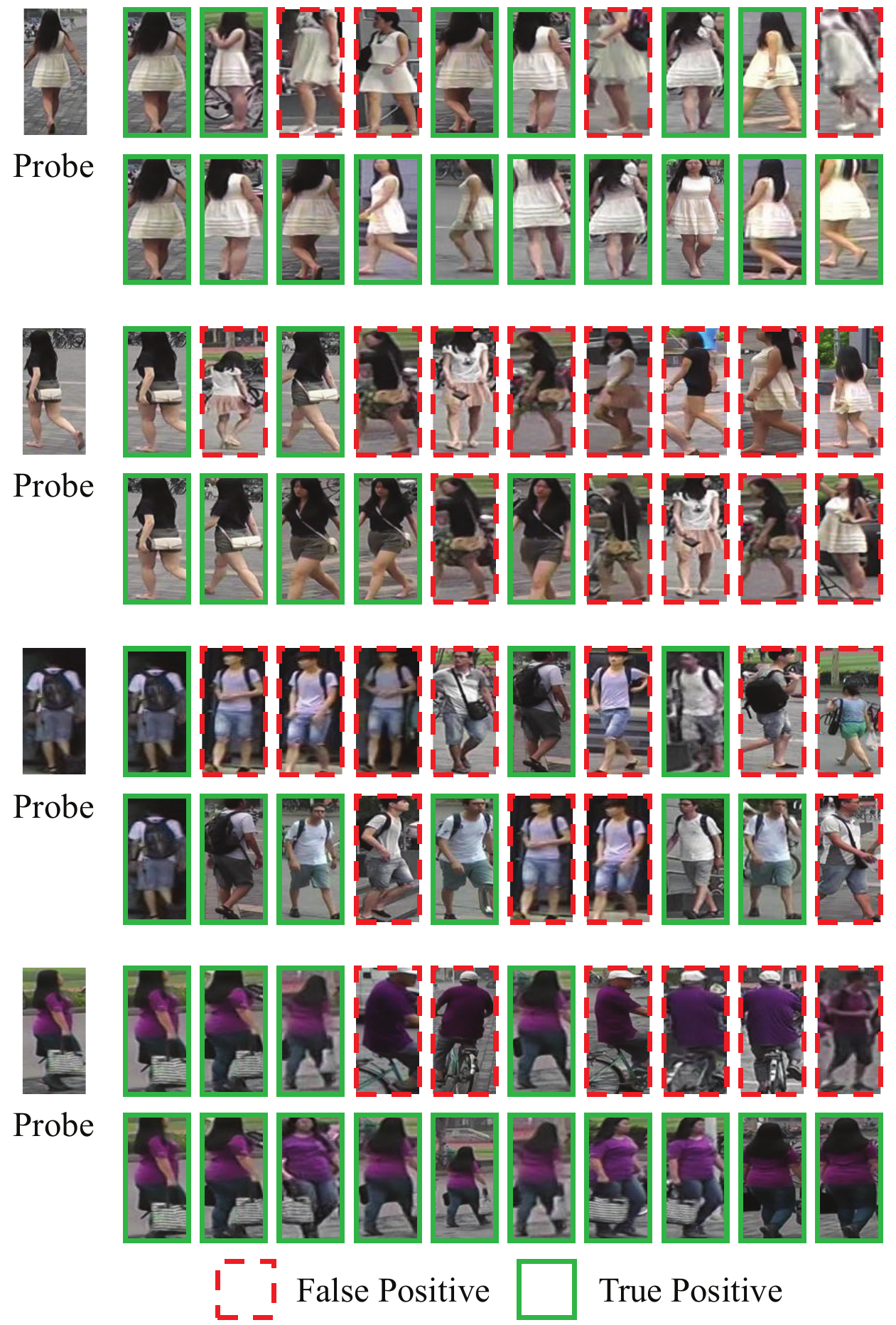}
\end{center}
\caption{Examples of Re-ID results on Market1501. In each example, the first row and second row show top-10 retrieved images of baseline and GLAD, respectively.}
\label{fig:Query}
\end{figure}

Our descriptor is learned by a four-stream neural network on the global and local regions. The four neural networks are trained with shared parameters in convolutional layers. We thus first compare different feature fusion and training strategies to test the validity of our descriptor learning method. The experimental results on Market1501 are shown in Table \ref{tab:GLN_Combination}. In Table \ref{tab:GLN_Combination}, \emph{WO/S} denotes training without parameter sharing, and \emph{W/S} denotes training with shared parameters. ``Global'' denotes $f^{G}$ extracted by our four stream network on the global image. ``Upper+Lower body'' fuses $f^{ub}$ and $ f^{lb}$. ``Head+Upper+Lower body'' fuses features on three local regions, and ``GLAD'' fuses the four descriptors in Eq.~\eqref{eq:Global-Local-D}.

In Table \ref{tab:GLN_Combination}, it is obvious that, sharing parameters during training substantially boosts the performance of learned descriptors. This might be because, the shared convolution kernels are forced to learn both global and local cues, thus are trained with more samples and could better avoid overfitting. We also observe that, the global feature performs better than features on local regions. This might be because the global image contains more visual cues thus reasonably conveys stronger discriminative power. This also explains why the discriminative power of descriptor on the head region is weaker than the ones on upper and lower body regions. As a result, ``Upper+Lower body'' outperforms ``Head+Upper+Lower body'',\emph{ i.e.}, equally fusing another descriptor with low discriminative power degrades the performance of $f^{ub}$ and $ f^{lb}$.  ``Head+Upper+Lower body (W)'' denotes fusing the three descriptors with different weights decided by the size of the three regions, \emph{ i.e.}, we set three weights as 0.2, 0.4, 0.4, respectively. It is can be observed that, weighted fusion results in better accuracy than both ``Upper+Lower body'' and ``Head+Upper+Lower body''. This means, with proper weight, the descriptor on head region is still helpful in improving the Re-ID performance.

By combining both the global and local descriptors, GLAD outperforms all of the global and fused local descriptors. For instance, GLAD significantly outperforms our baseline by $13.6\%$ on mAP and $9.2\%$ on Rank-1 accuracy. The above experiments show the validity of our descriptor learning strategy, \emph{i.e.}, embedding local and global cues in a four-stream network with shared convolutional layers. Examples of Re-ID results produced by our baseline and GLAD are shown in Fig. \ref{fig:Query}.

\subsection{ Evaluation on Part Extraction}
\label{sec:effect_of_part}

\begin{table}
\centering
\small
\caption{Performance of descriptors considering fine-grained part cues on Market1501. ``*'' denotes descriptors are extracted from the standard pose image~\cite{zheng2017pose} generated based on fine-grained part cues.}
\begin{tabular}{lccc}
\hline
 Training Strategy & Descriptor & mAP &  Rank-1\\
\hline
- & Baseline* &51.3 &72.6 \\
\hline
& Global* & $55.2$  & $76.8$ \\
& Upper+Lower body*   & $41.2$ & $68.9$ \\
W/S& Head+Upper+Lower body*    & $37.7$ & $67.4$ \\
& Head+Upper+Lower body (W)*     & $44.2$ & $72.0$ \\
& GLAD*   & $\bf{61.2}$ & $\bf{81.5}$\\
\hline
\end{tabular}
\label{tab:Fine_Grain}
\end{table}

Because it is difficult to detect accurate fine-grained parts on person Re-ID data, GLAD should be extracted on three coarse parts rather than the fine-grained parts. To estimate the validity of our part extraction strategy, the most intuitive way is to compare our descriptor with descriptors learned on fine-grained parts using the GLAD structure. Thus, we compare with a recent work that considers fine-grained part cues for descriptor learning~\cite{zheng2017pose}. In \cite{zheng2017pose}, 10 parts are captured and normalized with affine transform to achieve pose invariance. The normalized parts then compose a standard pose image, which is expected to be invariant to misalignment and pose changes. For fair comparison, we input the original image and the standard pose image generated by~\cite{zheng2017pose} into our descriptor learning module, then compare the learned GLAD descriptors on these two inputs. In other words, the two GLADs are learned with the same setting but on different inputs, \emph{i.e.}, our method embeds coarse part cues, and the other considers fine-grained parts and extra affine transformation. If the fine-grained part cues are helpful to learn robust descriptor, the GLAD considering fine-grained part cues should outperform the GLAD generated from coarse parts.

Table \ref{tab:GLN_Combination} shows the performance of GLAD extracted on coarse parts. The performance of GLAD considering fine-grained part cues is summarized in Table \ref{tab:Fine_Grain}. The comparison between Table \ref{tab:GLN_Combination} and Table \ref{tab:Fine_Grain} obviously shows that, descriptors generated on coarse parts gets better performance than those considering fine-grained parts cues. This conclusion thus supports our discussions, \emph{i.e.}, fine-grained part region detection is unstable and may degrade the performance of person Re-ID.

\begin{table}
\centering
\small
\caption{Comparison on Market1501 in single query mode.}
\begin{tabular}{lccc}
\hline
 Methods         &mAP     &Rank-1 \\
\hline
BoW+Kissme~\cite{market1501}       &20.8    &44.4\\
WARCA~\cite{WARCA}       &-    &45.2 \\
LOMO+XQDA~\cite{LOMOXQAD} &22.2    &43.8\\
Null Space~\cite{LDNS} &35.7  &61.0\\
SCSP~\cite{scsp} &26.4 &51.9 \\
\hline
PersonNet~\cite{wu2016personnet}     &26.4    &37.2 \\
Gated Siamese~\cite{GatedSiamese}    &39.6    &65.9 \\
LSTM Siamese~\cite{varior2016siamese}      &35.3    &61.6 \\
DLCNN~\cite{zheng2016discriminatively} &59.9 &79.5\\
PIE~\cite{zheng2017pose}     &56.0    &79.3 \\
\hline
Baseline & 60.3  & 80.7 \\
GLAD     &{\bf73.9}   &{\bf89.9}\\
\hline
\end{tabular}
\label{tab:market}
\end{table}

\subsection{Comparison with Other Methods}

To test the discriminative power of GLAD, we use the 4096-dim GLAD and squared Euclidean distance for person Re-ID. On market1501, we compare GLAD with many state-of-the-art works belonging to two categories, \emph{i.e.}, distance metric learning based methods and deep learning based methods, respectively. In the comparison, the metric learning based methods include Bow+Kissme \cite{market1501}, WARCA~\cite{WARCA}, LOMO+XQDA~\cite{LOMOXQAD}, Null Space \cite{LDNS}, SCSP \cite{scsp}. Deep learning based methods include Gated Siamese \cite{GatedSiamese}, LSTM Siamese \cite{varior2016siamese}, PersonNet \cite{wu2016personnet}, DLCNN \cite{zheng2016discriminatively} and PIE~\cite{zheng2017pose}. The results are shown in Table \ref{tab:market}. In the table, we observe that our method outperforms these previous works by large margins. For example, our method outperform the best result of those compared works by $10.4\%$ on Rank-1 accuracy and $14.0\%$ on mAP, respectively.

On CUHK03, we compare GLAD with recent distance metric learning based methods, including WARCA \cite{WARCA}, LOMO + XQDA \cite{LOMOXQAD}, Null Space \cite{LDNS} and MLAPG \cite{MLAPG}. Deep learning based methods including PersonNet \cite{wu2016personnet}, SI-CI \cite{SICI}, Gated Siamese \cite{GatedSiamese}, LSTM Siamese \cite{varior2016siamese}, Improved Deep \cite{ahmed2015improved}, DGD~\cite{xiao2016learning} and PIE \cite{zheng2017pose} are also compared. Experiments are conducted on both the datasets with labeled and detected bounding boxes. The results are show in Table \ref{tab:cuhk03lab} and Table \ref{tab:cuhk03dec}, respectively. From the two tables, it is clear that GLAD achieves promising performance. We achieve Rank-1 accuracy of $85.0\%$ on the labeled dataset and Rank-1 accuracy of $82.2\%$ on the detected dataset, which outperform all the other works.

The comparisons on VIPeR are summarized in Table \ref{tab:viper}. WARCA \cite{WARCA}, Null Space \cite{LDNS}, LOMO+XQDA \cite{LOMOXQAD}, Mirror-KMFA \cite{MirrorKFDA}, MLAPG \cite{MLAPG}, SCSP \cite{scsp} are compared as distance metric learning based methods. Deep learning based methods include Gated Siamese \cite{GatedSiamese}, LSTM Siamese \cite{varior2016siamese}, SI-CI \cite{SICI}, PIE \cite{zheng2017pose} and PIE+Mirror+MFA \cite{zheng2017pose}. We can observe that traditional distance metric learning based methods show substantial advantages over deep learning based methods. This is mainly because VIPeR is not large enough for deep model training. However, GLAD still achieves the best Rank-1 accuracy among all of those methods, and constantly outperforms all the other deep learning based methods at different rank levels.

It is also necessary to note that, PIE~\cite{zheng2017pose} considers fine-grained parts to learn global descriptors. Our method substantially outperforms PIE on the three datasets. This also shows the advantages of considering coarse part cues and explicitly embedding both local and global cues for descriptor learning.

\begin{table}
\centering
\small
\caption{Comparison on CUHK03 labeled dataset.}
\begin{tabular}{lccccc}
\hline
 Methods            &Rank-1   &Rank-5   &Rank-10   &Rank-20 \\
\hline
LOMO + XQDA~\cite{LOMOXQAD} &52.2    &82.2    &94.1    &96.3\\
WARCA~\cite{WARCA} &78.4    &94.6    &-   &-\\
MLAPG~\cite{MLAPG} &58.0 &87.1 &94.7 &96.9 \\
Null Space~\cite{LDNS}     &62.6    &90.1    &94.8    &98.1\\
\hline
PersonNet~\cite{wu2016personnet}            &64.8    &89.4    &94.9    &98.2 \\
Improved Deep~\cite{ahmed2015improved}                 &54.7    &86.5    &93.9    &98.1\\
DGD~\cite{xiao2016learning}                &72.6    &91.6    &95.2    &97.7\\
\hline
Baseline & 74.4  & 95.4 & 97.9  & 99.1 \\
GLAD      &{\bf85.0}&{\bf97.9}&{\bf99.1}&{\bf99.6}\\
\hline

\end{tabular}
\label{tab:cuhk03lab}
\end{table}

\begin{table}
\centering
\small
\caption{Comparison on CUHK03 detected dataset.}
\begin{tabular}{lccccc}
\hline
 Methods            &Rank-1    &Rank-5    &Rank-10  &Rank-20 \\
\hline
LOMO + XQDA~\cite{LOMOXQAD} &46.3    &78.9    &88.6    &94.3\\
MLAPG~\cite{MLAPG} &51.2 &83.6 &92.1 &96.9 \\
Null Space~\cite{LDNS}    &54.7    &84.8    &94.8    &95.2\\
\hline
SI-CI~\cite{SICI}                 &52.2    &84.3    &92.3    &95.0\\
Gated Siamese~\cite{GatedSiamese}                &61.8    &80.9    &88.3 &-\\
LSTM Siamese~\cite{varior2016siamese}                &57.3    &80.1    &88.3    &-\\
PIE~\cite{zheng2017pose}                &67.1    &92.2    &96.6    &98.1\\
\hline
Baseline & 70.4  & 93.3 & 97.0  & 98.7 \\
GLAD      &{\bf82.2}&{\bf95.8}&{\bf97.6}&{\bf98.7}\\
\hline
\end{tabular}
\label{tab:cuhk03dec}
\end{table}

\begin{table}
\centering
\small
\caption{Comparison on VIPeR dataset.}
\begin{tabular}{lccccc}
\hline
 Methods            &Rank-1    &Rank-5    &Rank-10   &Rank-20 \\
\hline
LOMO + XQDA~\cite{LOMOXQAD} &40.0    &67.4    &80.5    &91.1\\
WARCA~\cite{WARCA} &40.2   &68.2   &80.7    &91.1\\
Null Space~\cite{LDNS}    &51.2    &82.1    &90.5    &95.9\\
MLAPG~\cite{MLAPG} &40.7 &-  &82.3 &92.4 \\
Mirror-KMFA\cite{MirrorKFDA} &43.0 &75.8 &87.3 &94.8\\
SCSP~\cite{scsp} &53.5 &{\bf82.6} &{\bf91.5} &{\bf96.7} \\
\hline
SI-CI~\cite{SICI}                 &35.8    &67.4    &83.5    &-\\
Gated Siamese~\cite{GatedSiamese}                &37.8    &66.9    &77.4 &-\\
LSTM Siamese~\cite{varior2016siamese}                &42.4    &68.7    &79.4    &-\\
PIE~\cite{zheng2017pose}                &27.4    &43.0    &50.8    &60.2\\
PIE+Mirror+MFA~\cite{zheng2017pose}                &43.3    &69.4    &80.4    &89.9\\
\hline
Baseline & 39.2  & 63.3 & 75.6  & 82.9 \\
GLAD      &{\bf54.8}&{74.5}&{83.5}&{91.8}\\
\hline
\end{tabular}
\label{tab:viper}
\end{table}

\subsection{Performance of Retrieval Framework}
\label{sec:E_RS}

Market1501 allows to implement person Re-ID as a pedestrian retrieval task. Therefore, we use Market1501 to evaluate our retrieval framework. Our indexing and retrieval method involves one parameter $\theta$ in TDC, which is the threshold of dissimilarity degree within each group. We thus first test the impact of $\theta$ on Re-ID performance. Experimental results are summarized in Table~\ref{tab:TDC}.

Table~\ref{tab:TDC} shows different $\theta$ affect the number of generated groups. Smaller $\theta$ requires larger similarity within each group, thus divides the samples in gallery into more groups. Note that, the group number equals to the sample number when $\theta=0$. It is obvious that smaller $\theta$ improves the Re-ID performance. This is because smaller $\theta$ tends to exclude outliers in each group and produces more accurate coarse retrieval results. It also can been seen that, the final accuracy is mainly decided by the fine retrieval and is not sensitive to $\theta$. $\theta$ obviously affects the retrieval efficiency. Larger $\theta$ generates fewer groups,\emph{ i.e.}, produces smaller search space for coarse retrieval.

To further show the benefit of coarse retrieval, we test the impact of feature dimensionality in coarse retrieval in Table~\ref{tab:TDC}. With low dimensionality, the retrieval speed is substantially improved. For example, with $\theta=0.0015$, reducing the dimensionality from 4096 to 128 substantially reduces the retrieval time from 176ms to 31ms. The mAP and Rank-1 accuracy are almost unchanged, \emph{e.g.}, 73.2\% vs. 73\% on mAP and 89.9\% vs. 89.8\% on Rank-1 accuracy, respectively. Such accuracy is similar to the one of GLAD in Table~\ref{tab:market}. With 128 or 512-dim feature, increasing $\theta$ does not constantly improve the efficiency, because fine retrieval dominates the query time. Higher $\theta$ generates larger group size, and enlarges the search space of fine retrieval,\emph{ i.e.}, images in top-100 relevant groups.

Those comparisons imply our indexing and retrieval framework effectively accelerates the online Re-ID without degrading the accuracy. We thus conclude that, our proposed person Re-ID procedure is efficient and expected to work well in large-scale person Re-ID tasks.

\begin{table}
\centering
\small
\caption{Re-ID performance with different $\theta$ and feature dimension in coarse retrieval.}
\begin{tabular}{lcccccc}
\hline
$\theta$ & Group Number & Dim &mAP &  Rank-1 &Times(ms)\\
\hline
$0.0000$ & $19732$   & $4096$ & $73.9$  & $89.9$  &368\\
$0.0010$ & $13509$ & $4096$ & $73.7$  & $89.9$  &267\\
${\bf0.0015}$ & ${\bf8509}$   & ${\bf4096}$  & ${\bf73.2}$ & ${\bf89.9}$ &{\bf176}\\
$0.0020$ & $2558$  & $4096$   & $71.7$ & $89.8$ &101\\
\hline
$0.0015$ & $8509$ & $512$     & $73.1$  & $89.9$ &50 \\
$0.0015$ & $8509$ & $128$     & $73.0$  & $89.8$ &31 \\
$0.0020$ & $2558$ & $512$     & $71.6$  & $89.7$ &69 \\
$0.0020$ & $2558$ & $128$     & $71.4$  & $89.7$ &61 \\
\hline
\end{tabular}
\label{tab:TDC}
\end{table}

\section{Conclusion and Future Work}
\label{sec:conclusion}
This paper presents a GLAD descriptor and an efficient indexing and retrieval framework for pedestrian image retrieval. We first discuss that person Re-ID can be tackled in a retrieval task. GLAD is hence proposed with the motivation of generating a discriminative feature descriptor robust to misalignment and pose change issues. GLAD is extracted by explicitly learning the global and coarse part cues in human body through a four-stream CNN model. An efficient indexing and retrieval framework is finally proposed to accelerate the online Re-ID procedure. In this framework, the pedestrian images in gallery set are first divided into groups with TDC for offline indexing. Online retrieval first retrieves groups, then conducts fine retrieval to get a precise image ranklist. Extensive experiments show the strong discriminative power of GLAD and high speed of person Re-ID based on our indexing and retrieval framework.

Our current offline indexing needs to compute the pair-wise similarity between images, thus requires high offline complexity. Although this does not affect the online efficiency, more efficient strategies like hashing and approximate k-NN methods will be explored in our future work. Moreover, better offline grouping and group feature extraction strategies will be studied, \emph{e.g.}, considering extra features like time stamp and location cues.

\section*{ACKNOWLEDGMENTS}
This work is supported by National Science Foundation of China under Grant No. 61572050, 91538111, 61620106009, 61429201, and the National 1000 Youth Talents Plan, in part to Dr. Qi Tian by ARO grant W911NF-15-1-0290 and Faculty Research Gift Awards by NEC Laboratories of America and Blippar.
\newpage

\bibliographystyle{ACM-Reference-Format}
\balance
\bibliography{egbib}

\end{document}